\def\year{2022}\relax
\documentclass[letterpaper]{article} 
\usepackage{aaai22}  
\usepackage{times}  
\usepackage{helvet}  
\usepackage{courier}  
\usepackage[hyphens]{url}  
\usepackage{graphicx} 
\usepackage{natbib}  
\usepackage{caption} 
\DeclareCaptionStyle{ruled}{labelfont=normalfont,labelsep=colon,strut=off} 
\usepackage{algorithm}
\usepackage{algorithmic}

\usepackage{newfloat}
\usepackage{listings}
\floatstyle{ruled}
\newfloat{listing}{tb}{lst}{}
\floatname{listing}{Listing}
\usepackage{amsmath}
\usepackage{amssymb}
\usepackage[capitalise]{cleveref}
\usepackage{booktabs}
\usepackage{multirow}
\usepackage{url}
\usepackage[per-mode=symbol-or-fraction]{siunitx}

\newcommand{\pmt}{PM\textsubscript{2.5}}

\newcommand{\notw}{NO\textsubscript{2}}
\newcommand{\notws}{NO\textsubscript{2} }

\newcommand{\ta}{\theta}
\newcommand{\ea}{\eta}
\newcommand{\D}{\mathcal{D}}
\newcommand{\E}{\mathbb{E}}
\newcommand{\Eop}{\mathop{\mathbb{E}}}
\newcommand{\V}{\mathbb{V}}
\newcommand{\N}{\mathcal{N}}
\newcommand{\eye}{\mathcal{I}}
\newcommand{\eabar}{\eta'}
\newcommand{\tabar}{\theta'}

\title{Bayesian Optimisation for Active Monitoring of Air Pollution}
\author{
    Sigrid Passano Hellan, Christopher G. Lucas and Nigel H. Goddard
}
\affiliations{
    School of Informatics\\
    University of Edinburgh, UK

    \{s.p.hellan, c.lucas, nigel.goddard\}@ed.ac.uk

}

\begin{document}

\maketitle

\begin{abstract}
Air pollution is one of the leading causes of mortality globally, resulting in millions of deaths each year. Efficient monitoring is important to measure exposure and enforce legal limits. New low-cost sensors can be deployed in greater numbers and in more varied locations, motivating the problem of efficient automated placement. Previous work suggests Bayesian optimisation is an appropriate method, but only considered a satellite data set, with data aggregated over all altitudes. It is ground-level pollution, that humans breathe, which matters most. We improve on those results using hierarchical models and evaluate our models on urban pollution data in London to show that Bayesian optimisation can be successfully applied to the problem.
\end{abstract}

\section{Introduction}
Ambient air pollution is one of the leading causes of death globally, with particulate matter with diameter less than 
\SI{2.5}{\micro\meter},
\pmt, causing 3-4 million deaths each year \cite{cohen_estimates_2017,lelieveld_contribution_2015}. 
One commonly monitored pollutant is nitrogen dioxide, \notw, as it is used to indicate the presence of traffic-generated air pollution \cite{katsouyanni_ambient_2003}. 
\notws is also part of an ozone-generating process, and short-term increases in ozone concentration are followed by short-term increases in mortality \cite{katsouyanni_ambient_2003}. Due to these adverse effects the WHO have produced guidelines limiting pollution concentrations, and many countries have adopted air pollution regulations \citep[p.~174-175]{world_health_organization_air_2006}.

Traditional air pollution monitoring uses large and expensive sensors that are typically managed by national or municipal authorities deciding where to locate sensors based on domain knowledge and constraints posed by the bulky nature of the sensors \cite{carminati_emerging_2017}. 
More recently, low-cost air pollution sensors have become available, e.g. \citet{liu_low-cost_2020} and \citet{kelly_ambient_2017}, which open up air pollution monitoring to more locations, and allow groups with limited budgets and domain expertise to create sensor networks, including local governments in developing nations as well as community groups around the world.
By automating the decision-making process for placing air pollution sensors, we can help these groups make the most of their limited resources.

This goal, to simplify the process for setting up new monitoring networks in a low-cost way, motivates our decision to focus on computationally relatively inexpensive models and simple feature sets.
Although models are generally improved by using data on road traffic and other pollution sources, these data can be difficult or expensive to obtain for developing nations or citizen scientists. We demonstrate that even a minimal feature set containing only sensor locations and readings can be useful.  Additionally, all the computations are done on CPUs, so access to GPUs is not needed.

Simplicity and explainability also guide the model design. Experimental exploration showed the need for informative priors, as simpler approaches without hyperparameter priors or with only basic priors did not perform well.
Instead, we found a pragmatic solution using related data to construct a prior through a hierarchical model. 
In that way data from other cities can be used for the prior, and the relevance of each other city inferred.
Gaussian processes are used to model the data, which have interpretable hyperparameters, aiding explainability. 
And the number of hyperparameters is limited by keeping the covariance functions simple. 

Bayesian optimisation (BO) has been used for pollution monitoring 
previously, but has tended either 
to estimate hyperparameters or their distributions in ways that do not 
generalise to new sites in a sample-efficient way 
\cite{ainslie_application_2009}
or to focus on robot trajectories 
\cite{morere_sequential_2017,singh_modeling_2010,marchant_bayesian_2012},
a distinct problem from the one we consider here.
Work has also been done on monitoring gas leaks
\cite{reggente_using_2009,asenov_active_2019}, 
but again by using a single moving agent. 
An exception is \citet{hellan_optimising_2020_arxiv},
but it is limited to satellite data as a proxy for ground measurements. 
We extend on it
with a more principled methodology and better results, as well as an 
evaluation on a more relevant data set taken from the London Air Quality 
Network (LAQN) \cite{imperial_college_london_london_1993_url}.
Our motivations are similar to \citet{smith_machine_2019_arxiv}, who use 
Gaussian processes for calibrating low-cost pollution sensors, 
but do not address the problem of sensor placement.

The main contribution of this paper is showing Bayesian optimisation to be useful for automating planning of pollution sensor networks. Specifically, we consider the problem of iteratively placing stationary sensors and locating the maximum average pollution in the given area.
Knowledge of the maximum lets us know whether regulations are being adhered to.
We also discuss the general usefulness of the methodology adopted.

\section{Background}

\subsection{Bayesian optimisation}

Bayesian optimisation \cite{shahriari_taking_2015} is an optimisation method based on maintaining a probabilistic model $m(x)$ of the underlying problem $f(x)$. At each iteration the model is fitted to the collected data, the next sampling location chosen according to the acquisition function $a(m(x))$ and a sample collected. The acquisition function balances the exploration/exploitation trade-off.
At each iteration, the problem $\max_x a(m(x))$ is solved instead of $ \max_x f(x) $. While evaluating $f(x)$ requires taking a measurement, $a(m(x))$ can be calculated from the model.
One acquisition function is expected improvement  \cite{jones_efficient_1998}, $a(x)=~\E [ \max(0, f(x)-f') |\: \D ]$ where $\D$ is the data observed and $f'$ is the highest sample from previous iterations.
The model is usually a Gaussian process (GP) \cite{rasmussen_gaussian_2006}.

\subsection{Hierarchical Bayesian modelling}

Hierarchical Bayesian modelling is the principle of having layers of random variables built on top of each other \cite{shiffrin_survey_2008}. 
The advantage of using a hierarchical model for BO is that it can be used to transfer information about the distribution of GP hyperparameters across different contexts -- such as cities -- without assuming that these contexts are identical.
This is advantageous in applications like pollution monitoring where there are plentiful observations for some contexts but not others, e.g. cities with and without extensive pollution monitoring programs, and sample efficiency is paramount as each sample is expensive to collect. Hierarchical Bayesian modelling has been used for air pollution monitoring and modelling, e.g. \citet{dawkins_where_2020}, \citet{cocchi_hierarchical_2007} and \citet{sahu_hierarchical_2012}, but not, to our knowledge, in combination with Bayesian optimisation.

\subsection{Approximate inference}

\newcommand{\Dj}{\mathcal{D}_j}
\newcommand{\taj}{\theta_j}
\newcommand{\yt}{Y^*}
\newcommand{\xt}{X^*}

Although the standard approach for BO is to use a single point estimate for hyperparameters, a more Bayesian approach has several advantages, including better sample efficiency and more accurate estimates of posterior uncertainty \cite{de_ath_how_2021_arxiv,snoek_practical_2012}.
Instead of a single sample, a set of samples (Monte Carlo approximation) or a parameterised distribution (variational inference) can be used to represent the hyperparameters.
However, it makes the inference more computationally challenging.
The posterior can be calculated in closed form for the standard GP model with a point estimate of the hyperparameters \cite{rasmussen_gaussian_2006}, but this is not possible for the hierarchical models. 
Following previous work, we use Monte Carlo approximation rather than variational inference, as \citet{de_ath_how_2021_arxiv} found the latter to be less accurate. 
The Bayesian treatment of hyperparameters has been considered more widely for non-BO Gaussian processes,
e.g. in \citet{lalchand_approximate_2020}, \citet{murray_slice_2010} and \citet{williams1995gaussian}.

In general when using approximate inference for model hyperparameters, one tries to solve \cref{eq-approx-inf-standard}, where $\xt$ and $\yt$ are the test features and values, $\taj$ the model hyperparameters and $\Dj$ the observed data. In contrast, we are solving  \cref{eq-approx-inf-us} where $\D$ is observed data from related problems, i.e. our prior information. By factoring out the dependence on $\D$, we can split the inference problem into two separate ones, which we can solve separately. 
The first one takes into account $\D$ and only needs to be solved once using Markov chain Monte Carlo.
The second one takes into account $\Dj$ and is solved each time a new observation is added through the BO iterations, using importance weighting.
\begin{align}
     p(\yt|\xt,\Dj) &= \int p(\yt|\xt,\taj)p(\taj|\Dj)d\taj \label{eq-approx-inf-standard} \\
     p(\yt|\xt,\D,\Dj) 
    &= \int p(\yt|\xt,\taj)p(\taj|\Dj,\D)d\taj  \label{eq-approx-inf-us}
\end{align}
We found a simple Metropolis-Hastings based approach \cite{chib_understanding_1995} worked well, but gradient-aware methods, e.g. \citet{hoffman_no-u-turn_2014}, may be more appropriate for more complex kernels and higher-dimensional parameter spaces.

\section{Methodology}

While the heart of our method is a standard Bayesian optimisation loop, its effectiveness relies on problem-specific design decisions, data pre-processing, and pre-training in order to obtain an appropriate joint prior over GP parameters.

Pollutant concentrations tend to follow a log-normal distribution \cite{cats_prediction_1980}, so we log-transformed our concentration data to match the distributional assumptions behind Gaussian processes. We also standardised (satellite) or mean-centred (LAQN) our data, see the appendix. For evaluation, the observed data is considered to be the ground truth, so the variance of the noise in the GP likelihood is clamped to a small value (1e-06). 
The acquisition function used is expected improvement \cite{jones_efficient_1998} 
and importance weighting is used to adapt the hyperparameter samples to the problem at hand.
 Each problem is initiated with some randomly selected samples; 10 for the satellite data and 5 for the London data. The computing resources used are described in the appendix.

\subsection{GP models} \label{sec-method-gp}

The GP models are constructed from two base covariance functions, the RBF kernel $k_{\mathrm{R}}$ in \cref{eq-k-rbf} and a directed version $k_{\mathrm{W}}$ from \citet{hellan_optimising_2020_arxiv}, given in \cref{eq-k-dir}. 
The latter only considers the spatial distance orthogonal to a reference direction $\gamma$, conceptualised as the wind direction. $\tau$ is the difference between feature vectors.
In \cref{eq-k-dir}, $A~=~\begin{bmatrix}{} \sin(\gamma)^2 &  - \sin(\gamma)\cos(\gamma)  \\ - \sin(\gamma)\cos(\gamma) & \cos(\gamma)^2 \end{bmatrix} $.

\begin{align}
     k_{\mathrm{R},i}(\tau) &= \sigma_{\mathrm{r}, i}^2 \exp \left( \frac{-\tau^T\tau}{ l_{\mathrm{r},i}^2} \right) \label{eq-k-rbf}\\
    k_{\mathrm{W},i}(\tau)  &=\sigma_{\mathrm{w}, i}^2 \exp \left( \frac{-\tau^T A \tau}{ l_{\mathrm{w},i}^2} \right)  \label{eq-k-dir}
\end{align}

From $ k_{\mathrm{R}}$ and $k_{\mathrm{W}}$ three composite covariance functions are constructed, which are the ones evaluated in this paper. They are given in \cref{eq-k-rbf-rbf,eq-k-rbf-product,eq-k-rbf-directed} and consist of a sum of two terms, which are meant to capture the slowly-varying and faster-varying parts of the observed signal.
Including the noise hyperparameter, the RBF-RBF model has 5 hyperparameters, the RBF-Product model 7 and the Sum model 6.
\begin{align}
    k_{\mathrm{RBF-RBF}}(\tau) &= k_{\mathrm{R},1}(\tau) + k_{\mathrm{R},2}(\tau) \label{eq-k-rbf-rbf}\\
    k_{\mathrm{RBF-Product}}(\tau) &= k_{\mathrm{R},1}(\tau) + k_{\mathrm{R},2}(\tau) k_{\mathrm{W},3}(\tau) \label{eq-k-rbf-product} \\
    k_{\mathrm{Sum}}(\tau) &= k_{\mathrm{R},1}(\tau) + k_{\mathrm{W},2}(\tau) \label{eq-k-rbf-directed}
\end{align}

\subsection{Hierarchical structure} \label{sec-method-hier}

\begin{figure}[H]
  \centering
  \includegraphics[width=0.4 \textwidth]{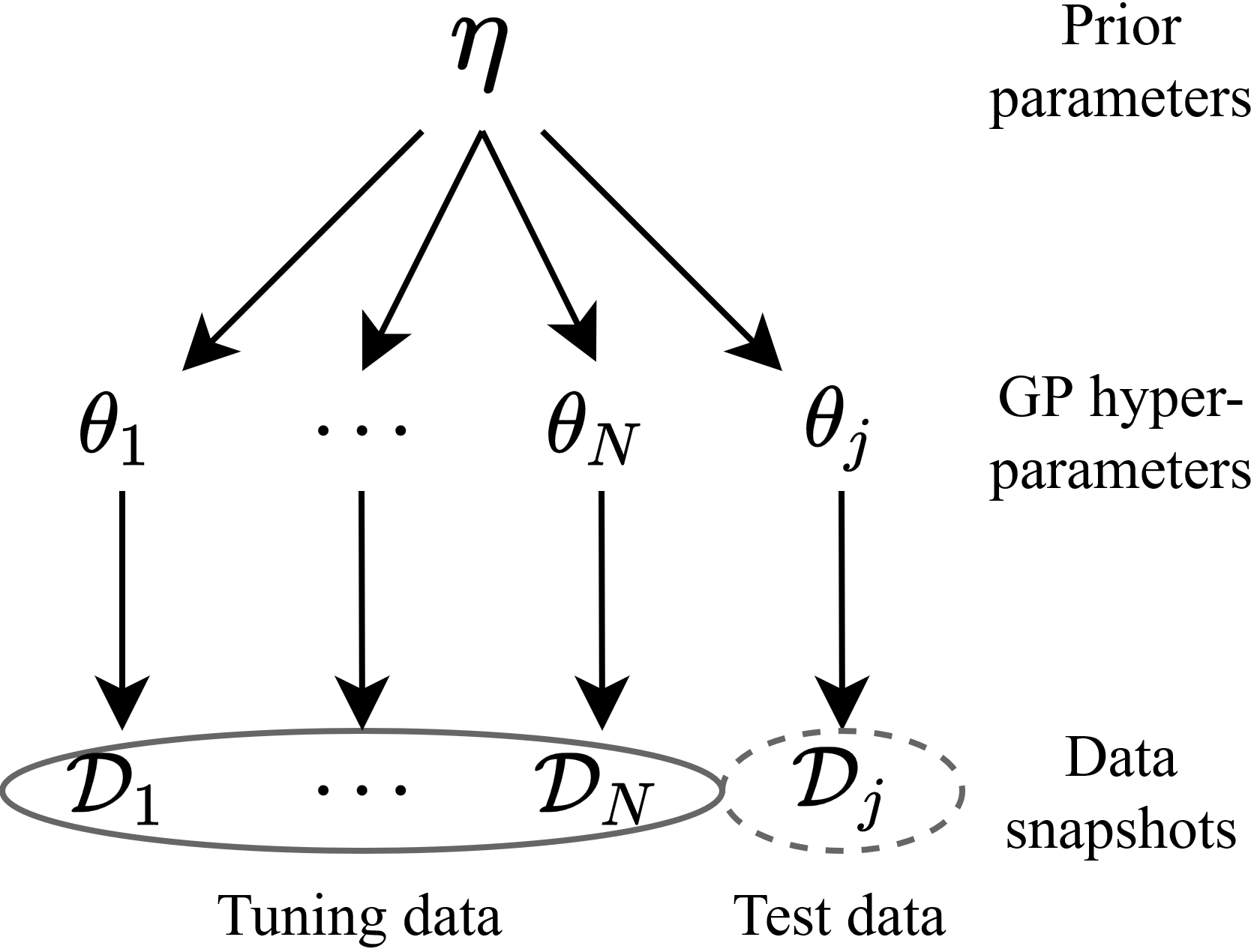}   
  \caption{Visualisation of the hierarchical structure adopted. The bottom level is the observed data $\D$. It is modelled using GPs, which are defined by their hyperparameters $\ta$. The distribution of the GP hyperparameters is captured by $\ea$. The hyperparameters $\ta$ are independent given $\ea$.}
  \label{fig:hierarchical-structure}
\end{figure}

Problem-appropriate hyperparameter settings are essential for Bayesian optimisation to be efficient. Too large a lengthscale $l_{\mathrm{r},i}$ and the process is assumed to vary too slowly and relevant locations are not explored. Too small a lengthscale and resources are wasted checking locations that are unlikely to be high and provide little new information. 
These pitfalls can be avoided using a prior that is informed by relevant data from other contexts. To that end, our model is structured hierarchically as described in \cref{fig:hierarchical-structure}, and the prior inferred from a related tuning set $(\D_1, \cdots, \D_N)$. 
The lowest level of the hierarchical model is the data snapshots, which are the only variables that are observed directly. GP models are fitted to the data snapshots, and the hyperparameters $\ta$ of the GP models constitute the middle layer of the model. 
The top level is the parameters defining the distribution of the GP hyperparameters. 
This can be expressed mathematically as 
$\displaystyle p(\ea,\ta,\D)\!~=~\!p(\ea)p(\ta|\ea)p(\D|\ta)\!~=~\!p(\ea)\prod_ip(\ta_i|\ea)p(\D_i|\ta_i) $.

The structure reflects the assumption that the data snapshots are generated by distinct but similar underlying processes.
Therefore, the tuning hyperparameters $(\ta_1, \cdots, \ta_N)$ cannot be used directly, but their distribution $\ea$ can be. 

For the BO loop we want the expectation of the acquisition function on snapshot $j$ at point $x_*$ given the distribution of GP hyperparameters, 
It can be approximated as 
\begin{align}
    a_j(x_*) &= \Eop_{p(\ta_i|\D,\Dj)} [a_{j,i}(x_*)] \\
     &= \Eop_{p(\ta_i|\ea)p(\Dj|\ta_i)/Z} [a_{j,i}(x_*)] \\
    & \approx \frac{1}{M} \sum_{i=1}^M a_{j,i}(x_*), \:\: \ta_i \sim \frac{1}{Z}p(\ta_i|\ea) p(\Dj|\ta_i) \\
     &\approx \frac{1}{W} \sum_{i=1}^M  a_{j,i}(x_*)p(\Dj|\ta_i), \:\: \ta_i \sim p(\ta_i|\ea) \label{eq-approx-samples} 
\end{align}
where $W = \sum_{i=1}^M  p(\Dj|\ta_i)$ is the sum of the importance weights and $Z$ a normalising constant. $a_{j,i}(x_*)$ is the acquisition function evaluated on snapshot $j$ using hyperparameter sample $\ta_i$ on the point $x_*$.
In the second line $\ea$ is used to summarise $\D$. In the third the expectation is approximated using Monte Carlo. In the fourth importance weighting is used so samples can be generated from the prior $\ea$, and weighted by their fit to the specific problem $\Dj$.
 By taking MCMC samples from $p(\ta_i|\ea)$ and not $p(\ta_i|\ea) p(\Dj|\ta_i)$ the computations are sped up greatly because 
 we can approximate the distribution once and then reuse it at each iteration of the BO loop and for each test data snapshot.

\subsection{MCMC sampling} \label{sec-method-mcmc}

The hyperparameter samples used for the prior are inferred from the tuning set using MCMC. It is done separately for each of the model kernels (RBF-Product, Sum and RBF-RBF). Joint samples are collected of the top two levels of the hierarchical structure from \cref{fig:hierarchical-structure}; $\ea$ and $\ta = \bigcup_{n \in N} \ta_n$. Note that $\taj$ is not included. $\ta_n$ is the hyperparameters for the GP model of $\D_n$, e.g. $(\smash{\sigma^2}_{\mathrm{r}, 1,n},l_{\mathrm{r},1,n},\smash{\sigma^2}_{\mathrm{r}, 2,n},l_{\mathrm{r},2,n})$ for the RBF-RBF kernel. Except for $\gamma$, each hyperparameter $\ta_{n,k}$, e.g. $l_{\mathrm{r},1}$, is assumed to be from a gamma distribution, $\ta_{n,k} \sim \Gamma (\psi_k,\phi_k)$, and $\ea$ is the set of parameters of those gamma distributions, $\ea = \bigcup_{k\in K} \{  \psi_k, \phi_k \}$. 
Thus $ \ta_n = \bigcup_{k \in K} \{ \ta_{n,k} \}$.

The sampling is done by alternatingly sampling $\ea$ and $\ta$. 
The tuning set $\{ \D_1, \ldots, \D_N\}$ is used to calculate the likelihood of the samples. 
The details of this sampling process are given in the appendix. 
The result is $H$ samples of $(\ea^{(h)},\ta^{(h)})$. The $M$ samples of $\ta_i$ for the importance weighting are generated by doing the following $M$ times: select $h$ at random, sample $\ta_{i,k} \sim \Gamma (\psi_{k}^{(h)},\phi_{k}^{(h)})$ for each $k$.
We use $M$=100 and $H$=1200 (satellite) or $H$=2000 (London), with a burn-in of 200 samples.

\section{Data}

\begin{figure}[h]
  \centering
  \includegraphics[width= 0.45\textwidth]{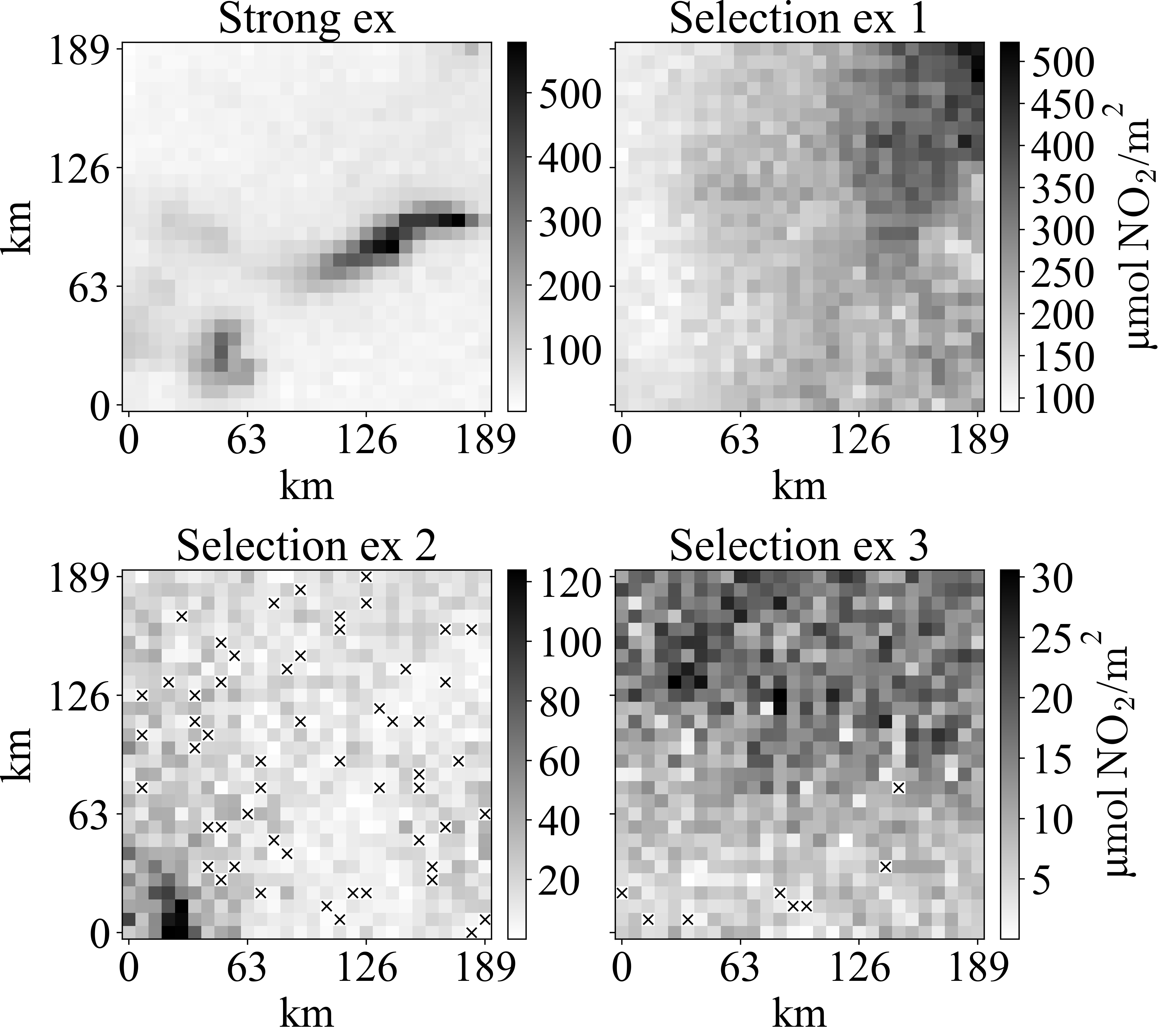}   
  \caption{Examples of data snapshots from the satellite data set. Selection examples 1, 2 and 3 are the strongest, median and weakest images from the selection subset, respectively. 
The strong example is adapted from \citet{hellan_optimising_2020_arxiv}. Crosses indicate missing data.}
  \label{fig:satellite-example}
\end{figure}

Two data sets are used for evaluation. The first comes from the TROPOspheric Measuring Instrument (TROPOMI) aboard the Sentinel-5P satellite from the EU's Copernicus programme \citep{copernicus_short}. The data set consists of 1083 images of 28x28 pixels, each giving the \notws concentration in $\mathrm{mol/m^2}$ within an area of about 7x7~km from the ground to the upper troposphere. The images are from October and November 2018, and have been selected for higher pollution concentrations. Images with more than 10~\% missing data were excluded.
The \notws concentrations vary greatly between images, so different subsets were considered. We use the same division as in \citet{hellan_optimising_2020_arxiv}. The `Strong' subset consists of the 50 images with the highest maxima, the `Median' subset of the 50 images with the median maxima and the `Weak' subset of the 50 images with the lowest maxima. Tuning sets consist of 10 images adjacent to these. The `Selection' subset consists of 100 images selected representatively from the whole set, without overlapping the above sets. Fig. \ref{fig:satellite-example} shows example images. The median tuning set is used for the selection subset.
For this data set, a data snapshot $\D_n$ refers to one image.

\begin{figure}[H]
  \centering
  \includegraphics[]{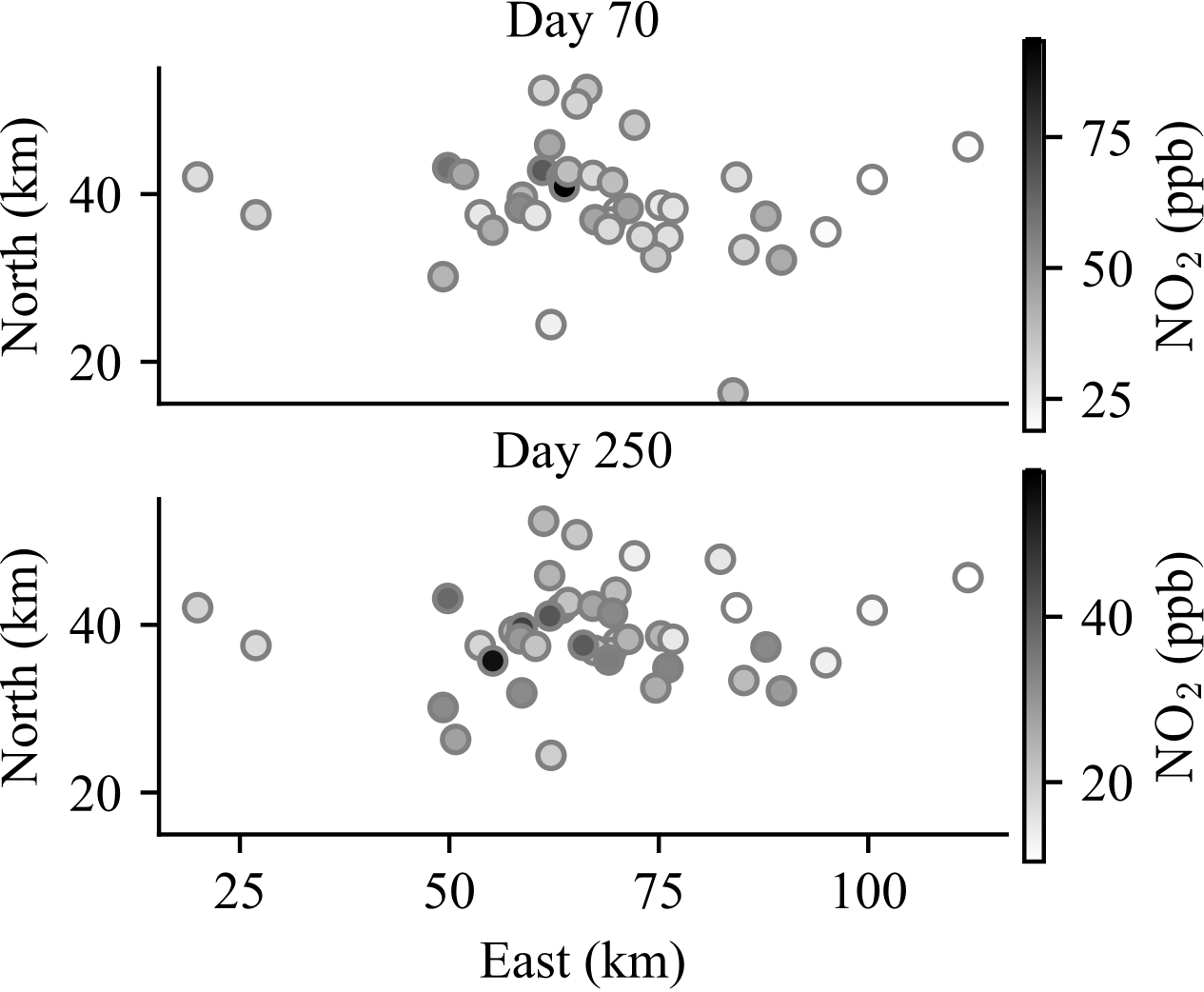}    
  \caption{Examples of data from the LAQN data set. The distances are from the most south-westerly monitoring station, in Beech outside Alton in Hampshire. Note that not all sensors are available each day, that the location of the maximum varies and the strong clustering in central London.}
  \label{fig:london-example}
\end{figure}

The second data set was extracted from the London Air Quality Network  \cite{imperial_college_london_london_1993_url}.
The full data set consists of years of data across multiple pollutants. For this paper, the \notws readings were used, and a tuning set constructed from the 2015 data and a test set from the 2016 data. Examples from the tuning set are shown in \cref{fig:london-example}. The days with less than 40 readings were discarded, and only the readings from `Roadside' sensors used. The former was done as days with more data are more suitable to show the benefits of BO. The latter was done as sensors with different  classifications behaved differently, so this simplified the modelling. Each day was used as a data snapshot $\D_n$. The tuning set consists of 214 such days and the test set 365 days.

\section{Results}
 
Two metrics are used to evaluate the model performance. Maximum ratio is the ratio between the true value at the estimated maximiser, $\hat{y}$, and the true maximum $y^*$. Since there are multiple data snapshots, the average is used, $R = \frac{1}{P} \sum_{p=1}^P \hat{y}_p/y^*_p$ where $P$ is the number of snapshots evaluated. The ratio metric is ideally one, with a higher score being better (it cannot exceed one).
The second metric is the Euclidean distance between the estimated maximiser $\hat{x}$ and the true maximiser $x^*$. Again, the average is used, $D = \frac{1}{P} \sum_{p=1}^P | \hat{x}_p-x^*_p |$. The distance metric is ideally zero, with a higher score being worse.
As in \citet{hellan_optimising_2020_arxiv}, we compare to a random baseline. 
At each iteration a location is chosen at random from those available. The result is given as the average over 100 runs on each of the data snapshots in each subset. For BO one run is used. The metrics are calculated on the pre-processed data, i.e. after log transform and standardisation / mean-centring.
 
\subsection{Satellite data}

\begin{figure}[H]
  \centering
  \includegraphics[]{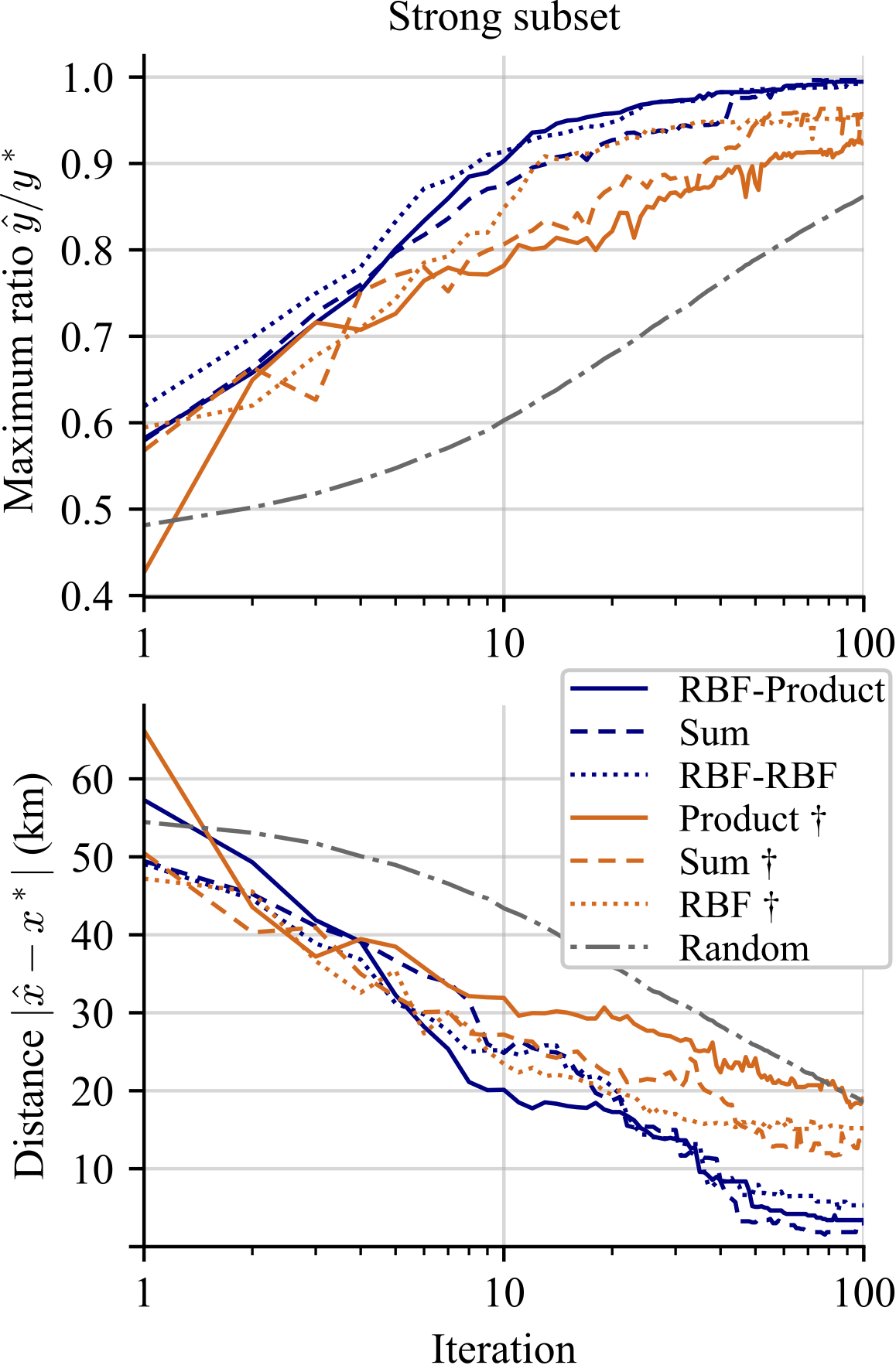}   
  \caption{Results on strong subset of satellite data. Our values in navy (darkest) compared to values in \citet{hellan_optimising_2020_arxiv} in orange (lightest) marked with $\dagger$. The baseline `Random' is shown in  grey (medium, different pattern). The results obtained are better than those in existing work, and both out-compete the random baseline. 
  $\hat{x}$ is the estimated maximiser and $x^*$ the true maximiser. $\hat{y}$ and $y^*$ are the true concentration values at $\hat{x}$ and $x^*$, respectively. }
  \label{fig:satellite-strong-res}
\end{figure}

\begin{figure}[H]
  \centering
  \includegraphics[]{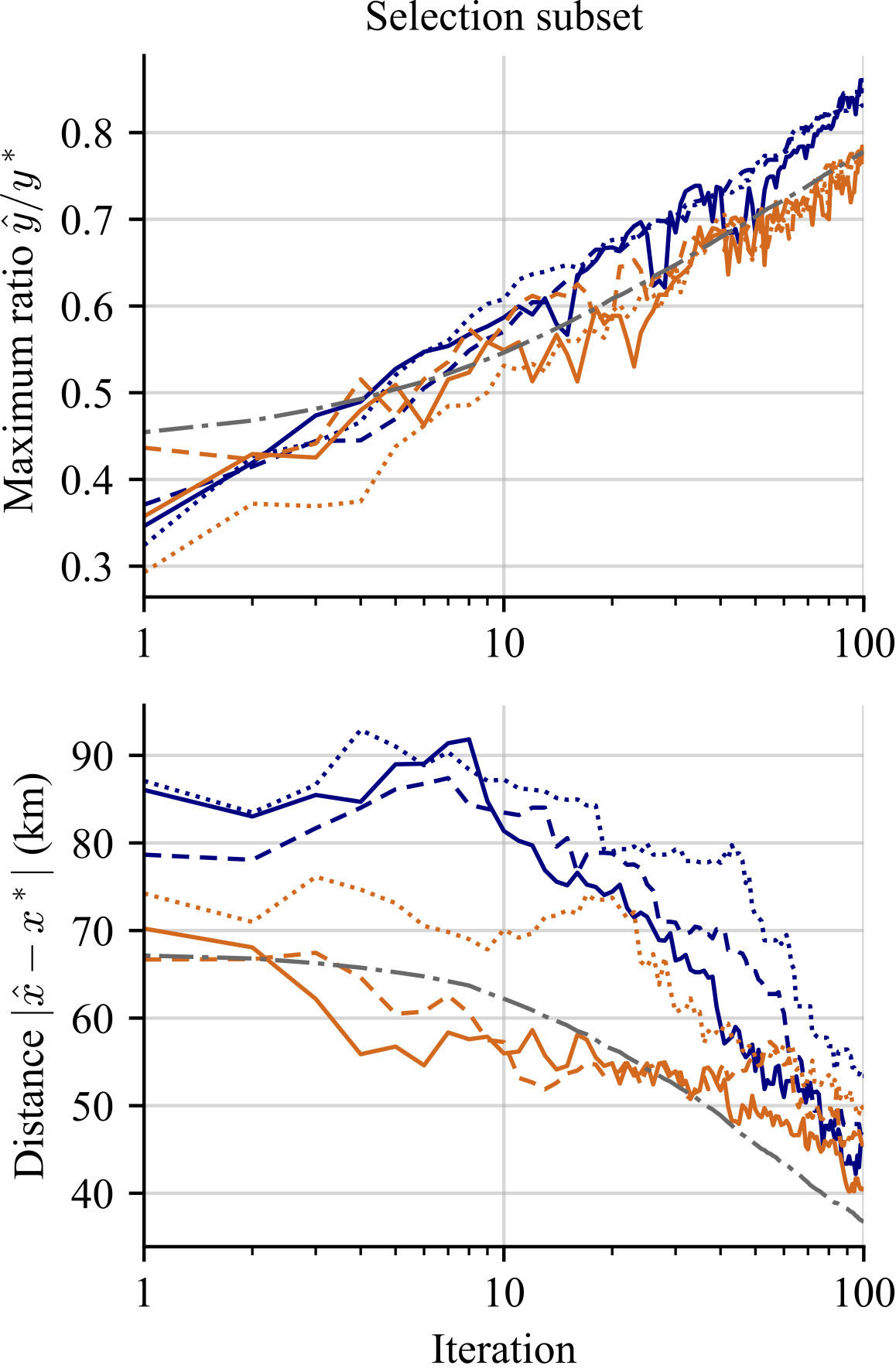}   
  \caption{Results on selection subset of satellite data. The legend and variable definitions are given in \cref{fig:satellite-strong-res}. Values in orange marked with $\dagger$ are from \citet{hellan_optimising_2020_arxiv}. The new results are an improvement on those in existing work when considering the ratio metric; 
   the three new results mostly lie on top of each other, and the three old results and the baseline mostly lie on top of each other. 
  The baseline is competitive on the distance metric.}
  \label{fig:satellite-selection-res}
\end{figure}

The results on the satellite data are given in \cref{fig:satellite-strong-res,fig:satellite-selection-res,tab:satellite-ratio-res,tab:satellite-dist-res}, which also give the results from \citet{hellan_optimising_2020_arxiv} for comparison. 
The performance has improved on both the strong and selection subsets.  
As can be seen, the previous work did not do better than the random guessing baseline on the selection subset.

\subsection{London data}

\begin{figure}[H]
  \centering
  \includegraphics[]{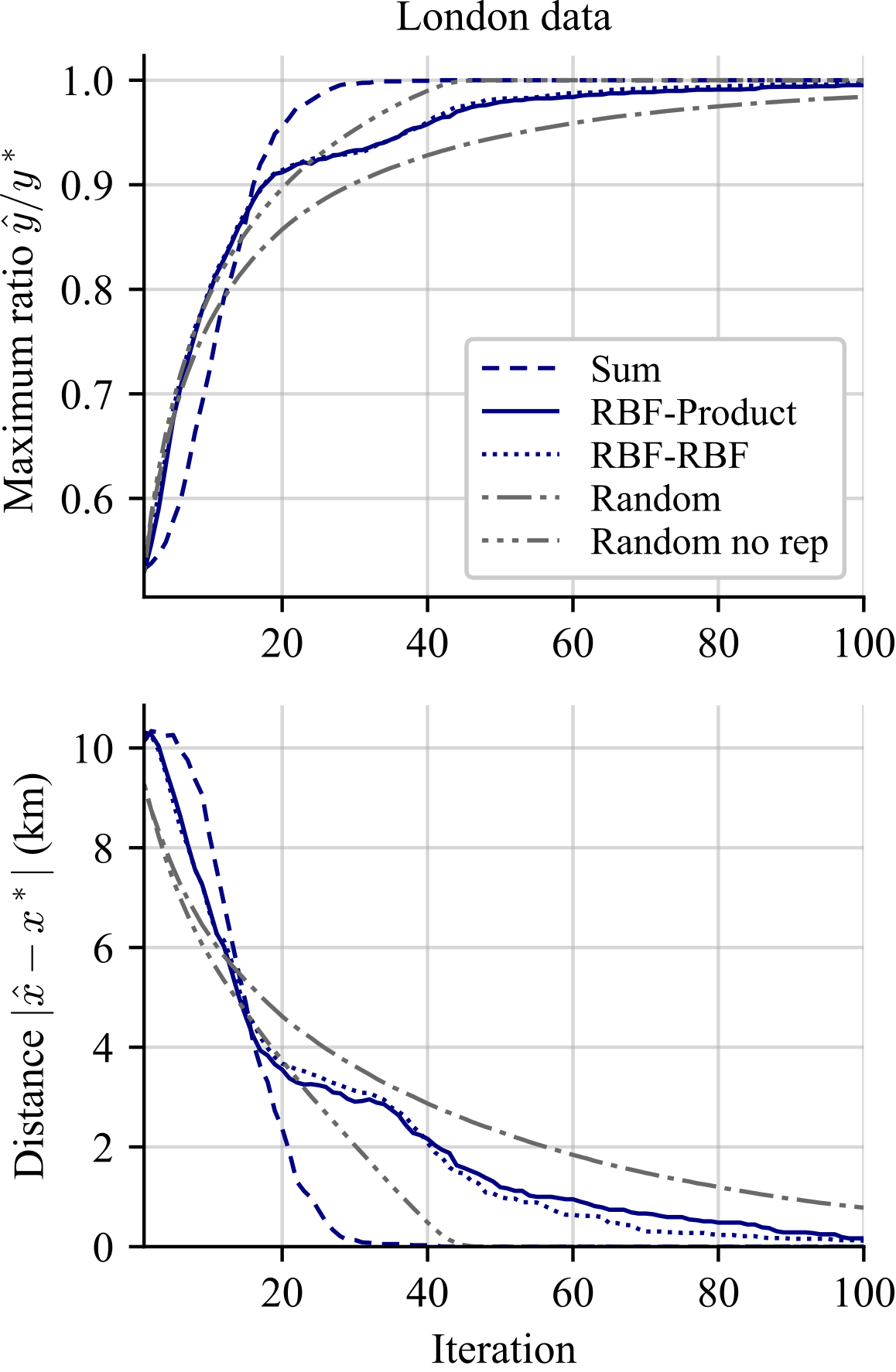}   
  \caption{Results on London data.  
  The variable definitions are given in \cref{fig:satellite-strong-res}. The RBF-Product and RBF-RBF lines follow each other closely and sometimes overlap.
  The Sum model is slower to start learning, but then finds the maximum much faster than the other models and the baselines.}
  \label{fig:london-res}
\end{figure}

The results on the London data are given in \cref{fig:london-res,tab:london-res}. As there are no previously published results for comparison, an additional random baseline is provided -- random selection without replacement. As all readings are assumed to be noise-free this leads to more efficient exploration. For the satellite data the difference between the baselines is small due to the larger number of available samples.

\begin{table}[H]
\centering
\caption{Confidence intervals for the means of the maximum ratio at the final iteration for the satellite data. Given are means $\pm$ one standard deviation of the mean. The best values are given in \textbf{bold}. This confirms the story from \cref{fig:satellite-strong-res,fig:satellite-selection-res} that  improved results are obtained on the strong and selection subsets. $\dagger$~indicates results from \citet{hellan_optimising_2020_arxiv}.}
\label{tab:satellite-ratio-res}
\begin{tabular}{@{}llll@{}}
\toprule
          & RBF-RBF     & Sum         & RBF-Product \\
Strong    & \textbf{0.987-0.997} & \textbf{0.993-0.999} & \textbf{0.990-0.999} \\ \cmidrule(l){2-4} 
          & RBF $\dagger$        & Sum $\dagger$        & Product $\dagger$    \\
          & 0.940-0.966 & 0.945-0.969 & 0.907-0.941 \\ \midrule
          & RBF-RBF     & Sum         & RBF-Product \\
Selection & \textbf{0.797-0.866} & \textbf{0.816-0.886} & \textbf{0.826-0.894} \\ \cmidrule(l){2-4} 
          & RBF $\dagger$       & Sum $\dagger$        & Product $\dagger$   \\
          & 0.755-0.804 & 0.756-0.802 & 0.750-0.810 \\ \bottomrule
\end{tabular}
\end{table}

\begin{table}[H]
\centering
\caption{Confidence intervals for the means of the maximiser distance at the final iteration for the satellite data. Given are means $\pm$ one standard deviation of the mean. The best values are given in \textbf{bold}. This confirms the story from \cref{fig:satellite-strong-res,fig:satellite-selection-res} that similar results are obtained on the selection subset and improved results on the strong subset. $\dagger$~indicates results from \citet{hellan_optimising_2020_arxiv}.}
\label{tab:satellite-dist-res}
\begin{tabular}{@{}llll@{}}
\toprule
          & RBF-RBF     & Sum         & RBF-Product \\
Strong    & \textbf{2.381-8.229} & \!\!\textbf{1.290-4.695} & \!\!\textbf{0.778-6.040} \\ \cmidrule(l){2-4} 
(km)          & RBF $\dagger$        & Sum $\dagger$        & Product $\dagger$    \\
          & 11.386-19.010 & \!\!8.445-16.976 & \!\!14.507-23.944 \\ \midrule
          & RBF-RBF     & Sum         & RBF-Product \\
Selection\!\! & 44.456-62.301 & \!\!36.960-56.178 & \!\!36.126-54.977 \\ \cmidrule(l){2-4} 
(km)        & RBF $\dagger$        & Sum $\dagger$        & Product $\dagger$    \\
          & 44.145-55.010 & \!\!40.551-51.918 & \!\!35.041-45.519 \\ \bottomrule
\end{tabular}
\end{table}

\begin{table}[ht]
\centering
\caption{
Confidence intervals for the means of the maximum ratio at the 31\textsuperscript{st} iteration for the London data. Given are means $\pm$ one standard deviation of the mean.
The best values are given in \textbf{bold}. 
}
\label{tab:london-res}
\begin{tabular}{@{}llll@{}}
\toprule
          & RBF-RBF     & Sum         & RBF-Product \\
Maximum    & 0.925-0.938 & \textbf{0.996-0.999} & 0.927-0.939 \\ \cmidrule(l){2-4} 
ratio          & Random       & Rand. no rep       &   \\
          & 0.904-0.905 & 0.957-0.958 &  \\ \midrule
          & RBF-RBF     & Sum         & RBF-Product \\
Distance & 2.841-3.330 & \textbf{0.042-0.133} & 2.694-3.148 \\ \cmidrule(l){2-4} 
(km)          & Random       & Rand. no rep       &   \\
          & 3.510-3.562 & 1.855-1.896 &   \\ \bottomrule
\end{tabular}
\end{table}

\section{Discussion}

The results on the satellite data, \cref{fig:satellite-strong-res,fig:satellite-selection-res,tab:satellite-ratio-res,tab:satellite-dist-res}, show that our method gives improved results even on the challenging selection subset.
The exception is the distance metric on the selection subset. 
An example of when a good ratio score corresponds to a bad distance score is an image that is zero everywhere except at the bottom right where it is 1.000, and at the top left where it is 0.999. 
The distance metric will be disproportionately high (bad) if only the marginally worse local optimum is found, but the ratio score will be good regardless. 
By inspection of \cref{fig:satellite-example} one can see that the selection examples have many local optima, and so would suffer from this.
On the London data one of the GP models does much better than either baseline, but the other two perform worse. Why this happens is examined in the Exploration subsection. But it shows that BO can be much faster, even if care must be taken when choosing the models and priors.
BO has the added benefit of providing models of the underlying process, with interpretable hyperparameters. An example of this is given in the Interpretability subsection.

Given the relatively small difference in performance between the GP models and the best random baseline one might be tempted to ask whether the extra effort needed for the former is justified. Therefore, it is worth keeping in mind the application. When deployed, each extra iteration will mean one extra sensor having to be procured and installed, requiring both money and labour. In that context, using more resources when planning the placement is preferable. For other domains, it becomes a trade-off between the cost of implementation and the cost of acquiring extra samples. If new samples can be acquired cheaply enough, then the random selection might be preferable. 

Throughout this paper, several simplifying assumptions have been made.
Firstly, although the London data is temporal, this has been simplified by constructing spatial problems and treating them as independent. 
Secondly, the measurements are treated as ground-truth readings. This simplifies the error metric calculations, but also makes irrelevant one of the advantages of probabilistically modelling the data, as this allows that uncertainty to be modelled directly. 
Another limitation is that the results on the satellite data are based on iteratively improving the method, and so could be considered validation results and not test results. The subsets are kept because there are published results to compare to. Because of how the data was split up it is not straightforward to generate a new Strong subset, but a new Selection subset was generated and the method evaluated on it. The results are given in the appendix, and show a strong improvement over the random baseline on both metrics.

\subsection{Exploration}

\begin{figure}[H]
  \centering
  \includegraphics[]{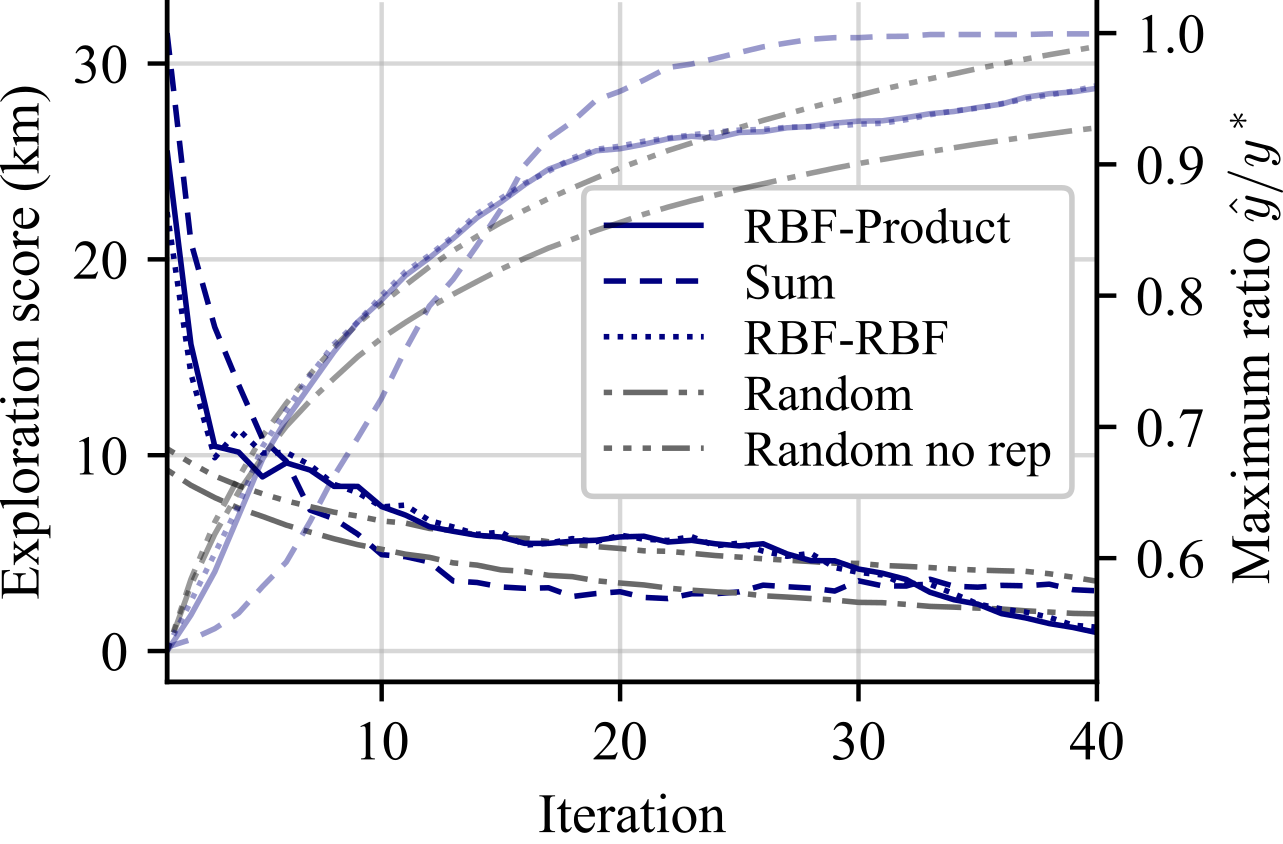}
  \caption{Exploration score of different methods on London data. 
  Overlaid and paler are the ratio scores from \cref{fig:london-res}. Note that Sum needs longer for the maximum ratio score to increase, and has a higher exploration score at the start.}
  \label{fig:analyse-exploration}
\end{figure}

The results in \cref{fig:london-res} can be examined in light of
the exploration-exploitation trade-off. 
An exploration score is used to do so, 
calculated as the minimum Euclidean distance between the new sample and any one of the already collected samples, $e_{i,j} = \min |x_{i,j} - x_{a,j}|, \:\:\: a<i$. This is done for each iteration $i$ and data snapshot $j$ and the average taken across the data 
snapshots; see \cref{fig:analyse-exploration}. 
It shows that the reason the Sum model improves more slowly in the beginning is that it emphasises exploration. Then, in the middle stage (5 to 20 iterations) it is able to learn faster, as it has already narrowed down what area to explore further, and manages to find the maximum faster. 
Conversely, and by chance, some of the random baselines' early iterations will be checking neighbouring points of those already seen, which can give small-scale improvement. 
The two other BO kernels start off with an exploration score between Sum and the first random baseline, and the performance is also between the two.

\subsection{Interpretability}

\begin{figure}[b]
  \centering
  \includegraphics[]{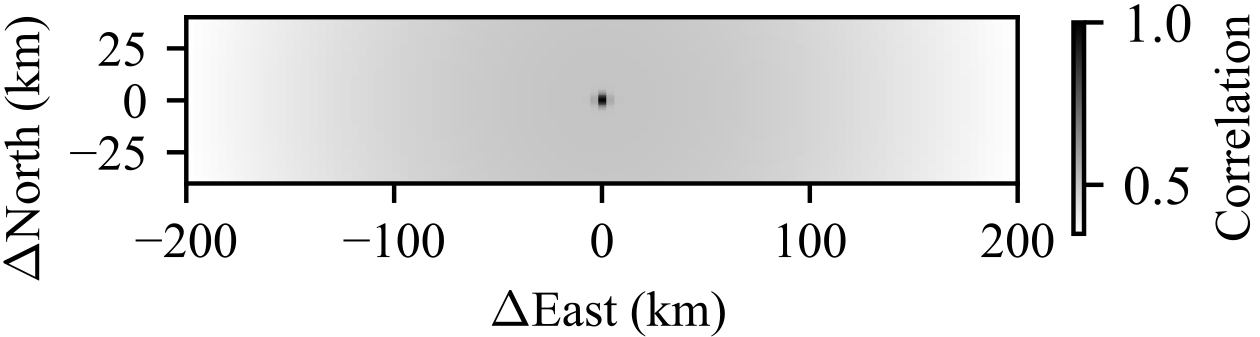}
  \caption{Expected correlation as function of distance for the mean of hyperparameter prior samples for the London data.}
  \label{fig:interpretability-vis}
\end{figure}

Part of our rationale for using comparatively simple GP kernels is to facilitate interpretation of hyperparameters, and analysis of the distribution patterns of pollution in an area, e.g. its smoothness and variability in space and time. This lets us perform ``at-a-glance" comparisons of different areas, discern whether extrema are more attributable to local sources or long-range patterns, and importantly lets domain experts assess the plausibility of a model's hyperparameters.

To illustrate, we consider the lengthscales $l_{\mathrm{r},i}$ of the RBF-RBF model in \cref{eq-k-rbf-rbf} on the London data.
 For simplicity, we take the mean of the samples generated for the prior, 
 as that shows the trends better than the posterior on a single day. Converting the values to the standard RBF expression 
gives
$l_{\mathrm{r},1}$~=~2.00~km, $l_{\mathrm{r},2}$~=~241~km, $\sigma^2_{\mathrm{r},1}$~=~2.05 and $\sigma^2_{\mathrm{r},2}$~=~2.04. 
This corresponds to one local component and one regional component  contributing roughly equally to the overall variability of \notws concentrations.
At a distance of 100 metres the expected correlation between \notws concentrations will be close to 1, driven by both local and long-range effects, whereas at a distance of 10~km it falls to 0.5, and local effects play a negligible role. The correlations are also visualised in \cref{fig:interpretability-vis}.
The LAQN provides a pollution map online at their website \cite{imperial_college_london_london_1993_url}. 
It shows pollution across all of London, but with more concentrated in central London radiating outwards and around Heathrow airport, as well as along major roads.
 This supports the modelling hypothesis of a local and a regional component.

\section{Conclusion}

We have shown that Bayesian optimisation with hierarchical models can be successfully applied to ground-level urban pollution data. 
We also presented a pragmatic method for approximate inference when some of the work can be precomputed, which we believe can be useful in other applications.
In Bayesian optimisation the GP tuning step needs to be repeated many more times than in standard regression, as new data points are added iteratively. Therefore, using standard MCMC on the full hierarchical model for each new data point would slow down the process. By using importance weighting in the BO tuning step the process can be sped up. This is even more useful in other
applications where collecting samples is less time consuming than for pollution.

As expected, the models performed worse on the ground-level data.
This is due in part to the data snapshots having fewer available samples. Some exploration is needed before the model can usefully discern the areas of interest, and where the data is sparse compared to the local variations this is harder.  
The strength of BO lies in being able to exploit structure in data, so it works better for slowly varying processes.
While the results show that the approach has promise, more evaluation is needed. In particular, due to data constraints, London was used both for developing the prior and for evaluation. 
To test whether the prior can be constructed from other cities additional data needs to be used. It would also be beneficial to test the methods as they would be used, i.e. iteratively placing sensors, but this would be much more resource-intensive. 
There might be easy performance gains available by experimenting with 
kernel families (e.g. Mat\'{e}rn) and combinations, 
such as more complex compositional kernels to model correlations at different scales. The benefits of modelling seasonal and weekly patterns should also be explored.
Additionally, the results showed the benefit of early exploration, so the methods could be biased towards this
by modifying the acquisition function.
\citet{berk_exploration_2018} present a modification to expected improvement that increases early exploration without needing manual tuning.

There are two main extensions needed before deploying the method: 
temporal modelling and accommodating sensors with varying precision and reliability. 
Temporal modelling is important as pollution levels vary temporally, with some days bringing high pollution levels throughout a city.
The temporal aspect can be avoided by only considering averages, but this ignores the legal limits on hourly averages  \citep[p.~174-175]{world_health_organization_air_2006}
and makes adding data from new sensors harder.
The uncertainty of a pollution reading is important when using low-cost sensors for monitoring to avoid misleading statements. \citet{smith_machine_2019_arxiv} present a solution to this, but do not discuss how to place the sensors. 
Unifying the active 
placement with the uncertainty treatment is necessary for 
successfully applying it using a network of low-cost sensors.

\section{Ethical considerations}

We believe the risks are very low from our proposed method. The only data used are \notws concentrations, and the latitude and longitude of the stationary sensors. No images or personal data of any kinds are used. This is an advantage over other proposed solutions, e.g. using taxis to collect data.
The potential benefits far outweigh the harms, by allowing local communities to keep authorities accountable. 

\section{Acknowledgements}

The authors would like to thank Dr Douglas Finch at the University of Edinburgh for extracting the
data and helpful correspondence, and Professor Iain Murray, also of the University of Edinburgh, for helpful advice and discussions.
The authors also thank the anonymous reviewers for their time and thoughtful comments. 

This work was supported by the EPSRC Centre for Doctoral Training in Data Science, funded
by the UK Engineering and Physical Sciences Research Council (grant EP/L016427/1) and the
University of Edinburgh.

\small
\bibliography{references,extra}

\normalsize
\newpage
\:
\newpage
\appendix

\section{Appendix}

HLG20 is here used to refer to \citet{hellan_optimising_2020_arxiv}.

\newcommand{\Kh}{K_j \left(\ta_i \right)}
\newcommand{\Khs}{K_{*,j} \left(\ta_i \right)}
\newcommand{\Khss}{K_{**} \left(\ta_i \right)}

\newcommand{\eanj}{\ea_{\not{j}}}

\newcommand{\tankh}{\ta_{n,k}^{(h)}}
\newcommand{\tankbar}{\tabar_{n,k}}
\newcommand{\tankhm}{\ta_{n,k}^{(h-1)}}
\newcommand{\eah}{\ea^{(h)}}
\newcommand{\eahm}{\ea^{(h-1)}}
\newcommand{\tanotk}{\ta_{n, \not{ k}}^{(h)}}
\newcommand{\eakgh}{\ea_{k,g}^{(h)}}
\newcommand{\eakghm}{\ea_{k,g}^{(h-1)}}
\newcommand{\eaknotg}{\ea_{k,\not g}^{(h)}}
\newcommand{\eakgbar}{\eabar_{k,g}}

\subsection{Acquisition function evaluation}

 At each BO iteration, importance weighting is used to calculate the acquisition function given in \cref{eq-exp_imp}. $a_{j,i}(x_*)$ is the acquisition function for test data snapshot $\D_j$ using hyperparameter sample $\ta_i$ on test point $x_*$. We use expected improvement \cite{jones_efficient_1998} as the base acquisition function. $f(x)$ is the underlying function and $f'$ the highest sample from previous iterations.
 \begin{align}
a_{j,i}(x_*) &= \E [ \max(0, f(x_*)-f') |\: \D_j, \ta_i ] \\
= \int_{f'}^{\infty}(f(x_*&)-f') \N(f;\mu_{j,i}(x_*),\V_{j,i}(x_*)) \mathrm{d}f \label{eq-exp_imp} 
\end{align}
The hyperparameter sample $\ta_i$ is used to calculate the posterior distribution, expressed by the mean and the variance in \cref{eq-gp-mean} and \cref{eq-gp-var}, respectively. $\Kh$ is the covariance of the observed points from $\D_j$, i.e. the ones sampled at previous iterations, using hyperparameter sample $\ta_i$. $Y_j$ is the values from the observed points. $\Khss$ is the covariance function evaluated on $\tau = x_* - x_* = \boldsymbol{0}$.
$\Khs$ is the covariance between the observed points and the test point.
$\sigma_n^2$ is the variance of the Gaussian noise in the likelihood.
\begin{align}
    \mu_{j,i}(x_*) &= \Khs^T (\Kh + \sigma_n^2 \eye)^{-1} Y_j \label{eq-gp-mean} \\
    \V_{j,i}(x_*) &=  \Khss \nonumber \\ 
    - &\Khs^T (\Kh + \sigma_n^2 \eye)^{-1} \Khs \label{eq-gp-var} 
\end{align}
Then, importance weighting is used to scale the acquisition function values from each hyperparameter sample. The weight is given in \cref{eq-gp-weight} and the final expression in \cref{eq-exp_imp-weighted}. $W = \sum_{i=1}^M  p(\Dj|\ta_i)$ as defined in the main paper.
\begin{align}
    w_{j,i} &= p(\D_j|\ta_i) = \N \left(\D_j;\boldsymbol{0}, \Kh \right) \label{eq-gp-weight} \\
    a_{j}(x_*)  &\approx W ^{-1} \sum_i w_{j,i}  a_{j,i}(x_*) \label{eq-exp_imp-weighted} 
\end{align}

\subsection{Details of MCMC sampling}

The sampling is done using MCMC to collect joint samples of $(\ea,\ta)$, where $\theta$ is the GP hyperparameters and $\ea$ the distribution relating $\theta$ for different problems, see \cref{fig:hierarchical-structure}. This is done using Metropolis-Hastings \cite{chib_understanding_1995} updating one variable $\psi_k$, $\phi_k$ or $\ta_{n,k}$ at a time.
The sampling is done according to \cref{algorithm-sampling}.

\begin{algorithm} 
\caption{Sampling $(\ea,\ta)$ } 
\label{algorithm-sampling} 
\begin{algorithmic} 
    \STATE Initialise $\ea^{(0)}$, $\ta^{(0)}$
    \FOR{$h$ = 1 $\ldots$ H}
        \FOR{$n$ = 1 $\ldots$ N}
            \FOR{$k$ = 1 $\ldots$ K}
                \STATE $\tanotk \Leftarrow \{
                \ta_{n,1}^{(h)}, \ldots, \ta_{n,k-1}^{(h)},\ta_{n,k+1}^{(h-1)}, \ldots, \ta_{n,K}^{(h-1)} \}$
                \STATE Sample $\tankh$ given $\eahm, \tanotk $
            \ENDFOR
        \ENDFOR
        \FOR{\_ = 1$\ldots$ B}
            \FOR{$k$ = 1 $\ldots$ K'}
                \FOR{$\ea_{k,g}^{(h)} = \psi_k^{(h)}, \phi_k^{(h)}$}
                    \STATE $\ea_{k,\not g}^{(h)} \Leftarrow \psi_k^{(h)} \:\:\mathrm{if}\:\: \ea_{k,g}^{(h)} \neq \psi_k^{(h)} \:\:\mathrm{else}\:\: \phi_k^{(h-1)}$
                    \STATE Sample $\ea_{k,g}^{(h)}$ given $\ta^{(h)},\ea_{k,\not g}^{(h)}$
                \ENDFOR
            \ENDFOR
        \ENDFOR
        \STATE Store sample $(\ea^{(h)},\ta^{(h)})$
    \ENDFOR
    \RETURN H samples of $(\ea,\ta)$
\end{algorithmic}
\end{algorithm}

The variables in \cref{algorithm-sampling} are defined as
\begin{itemize}
    \item H: Number of MCMC samples
    \item N: Number of tuning data snapshots
    \item K: Number of GP hyperparameters
    \item K': Number of GP hyperparameters with non-constant $\eta$
    \item B: Number of times to sample $\ea$ per $\ta$ sample
\end{itemize}
The $\eta$ parameters for the directional hyperparameter $\gamma$ and the noise hyperparameter are constant.

\subsubsection{Initialisation}

$\ea^{(0)}$ is set to a vector of ones.
$\ta^{(0)}$ is set by sampling from $\ea^{(0)}$.

\subsubsection{Sample $\tankh$ given $\eahm, \tanotk $}

The sampling of $\tankh$ given $\eahm$ and $\tanotk$ is done as follows.
At each iteration a proposal $\tankbar$ is generated, and either accepted or rejected proportionally to the acceptance ratio $a_\ta(\tankbar|\tankhm)$ where $\tankhm$ is the previous sample value. If the proposal is rejected the old value is used for the new sample.

$\tankh$ is the $k$-th GP hyperparameter fitted to data snapshot $\D_n$ at iteration $h$. The un-normalised probability function $\pi(\tankhm)$ is given in \cref{eq-ta-prob}. The proposal distribution is given in \cref{eq-ta-proposal} and the acceptance ratio in \cref{eq-ta-acceptance}.
$\pi(\tankbar)$ is evaluated by replacing $\tankhm$ in \cref{eq-ta-prob} with $\tankbar$. 
\begin{align}
    \pi(\tankhm) = p(\D_n&|\tankhm, \tanotk)p(\tankhm|\eahm)  \label{eq-ta-prob}\\
    Q_\ta (\tankhm|\tankbar) &= p(\tankhm|\eahm)  \label{eq-ta-proposal} \\
    a_\ta(\tankbar|\tankhm) &= \min \left( 1, \frac{  p(\D_n|\tankbar, \tanotk)}{  p(\D_n|\tankhm, \tanotk)} \right) \label{eq-ta-acceptance}
\end{align}
$p(\D_n|\tankhm, \tanotk)$ is evaluated through the standard GP expressions. $p(\tankhm|\eahm)=\Gamma (\tankhm;\psi_k^{(h-1)},\phi_k^{(h-1)})$ for
$k$ corresponding to $ \{\sigma^2_{\mathrm{r}}, l_{\mathrm{r}}, \sigma^2_{\mathrm{w}}, l_{\mathrm{w}}\}$ 
and $p(\tankhm|\eahm)= \mathrm{Uniform} (\tankhm;0,\pi)$ for $k$ corresponding to $ \gamma $.

\subsubsection{Sample $\ea_{k,g}^{(h)}$ given $\ta^{(h)},\ea_{k,\not g}^{(h)}$}

The sampling of $\ea_{k,g}^{(h)}$ given $\ta^{(h)}$ and $\ea_{k,\not g}^{(h)}$ is done as follows. At each iteration a proposal $\eakgbar$ is generated, and either accepted or rejected proportionally to the acceptance ratio $a_{\ea}(\eakgbar|\eakghm)$ where $\eakghm$ is the previous sample value. If the proposal is rejected the old value is used for the new sample.

The un-normalised probability function $\pi(\eakghm)$ is given in \cref{eq-ea-prob}. The proposal distribution is given in \cref{eq-ea-proposal} and the acceptance ratio in \cref{eq-ea-acceptance}. $\psi_k^{(h)}$ is the shape parameter of the gamma distribution, and $\phi_k^{(h)}$ the scale parameter.
The prior on $\eta$, $p(\eakghm)$, is set to one for valid values and zero otherwise.
\begin{align}
    \pi(\eakghm) =  &p(\eakghm) \prod_{n=1}^N p(\ta_{n,k}^{(h)}|\eakghm,\eaknotg) \label{eq-ea-prob}\\
    Q_{\ea} (\eakghm|\eakgbar) &= \N(\eakghm;\eakgbar, \sigma_{\ea,k,g}^2) \label{eq-ea-proposal} \\
    a_{\ea}(\eakgbar|\eakghm) &= \min \left( 1, \frac{\pi(\eakgbar)}{\pi(\eakghm) } \right) \label{eq-ea-acceptance}
\end{align}
$\sigma_{\ea,k,g}$ is a parameter determining the width of the proposal.
The values are 
\begin{itemize}
    \item $\sigma_{\ea,k,g}$=1.5 for $\eakghm$=$\psi_k^{(h)}$, $k$ corresponding to $l_{\mathrm{r}}$ or $l_{\mathrm{w}}$ 
    \item $\sigma_{\ea,k,g}$=0.5 for $\eakghm$=$\phi_k^{(h)}$, $k$ corresponding to $l_{\mathrm{r}}$ or $l_{\mathrm{w}}$ 
    \item $\sigma_{\ea,k,g}$=0.3 for $\eakghm$=$\psi_k^{(h)}$, $k$ corresponding to $\sigma^2_{\mathrm{r}}$ or $\sigma^2_{\mathrm{w}}$ 
    \item $\sigma_{\ea,k,g}$=0.1 for $\eakghm$=$\phi_k^{(h)}$, $k$ corresponding to $\sigma^2_{\mathrm{r}}$ or $\sigma^2_{\mathrm{w}}$ 
\end{itemize}

\subsection{Additional results}

\begin{figure}[ht]
  \centering
  \includegraphics[]{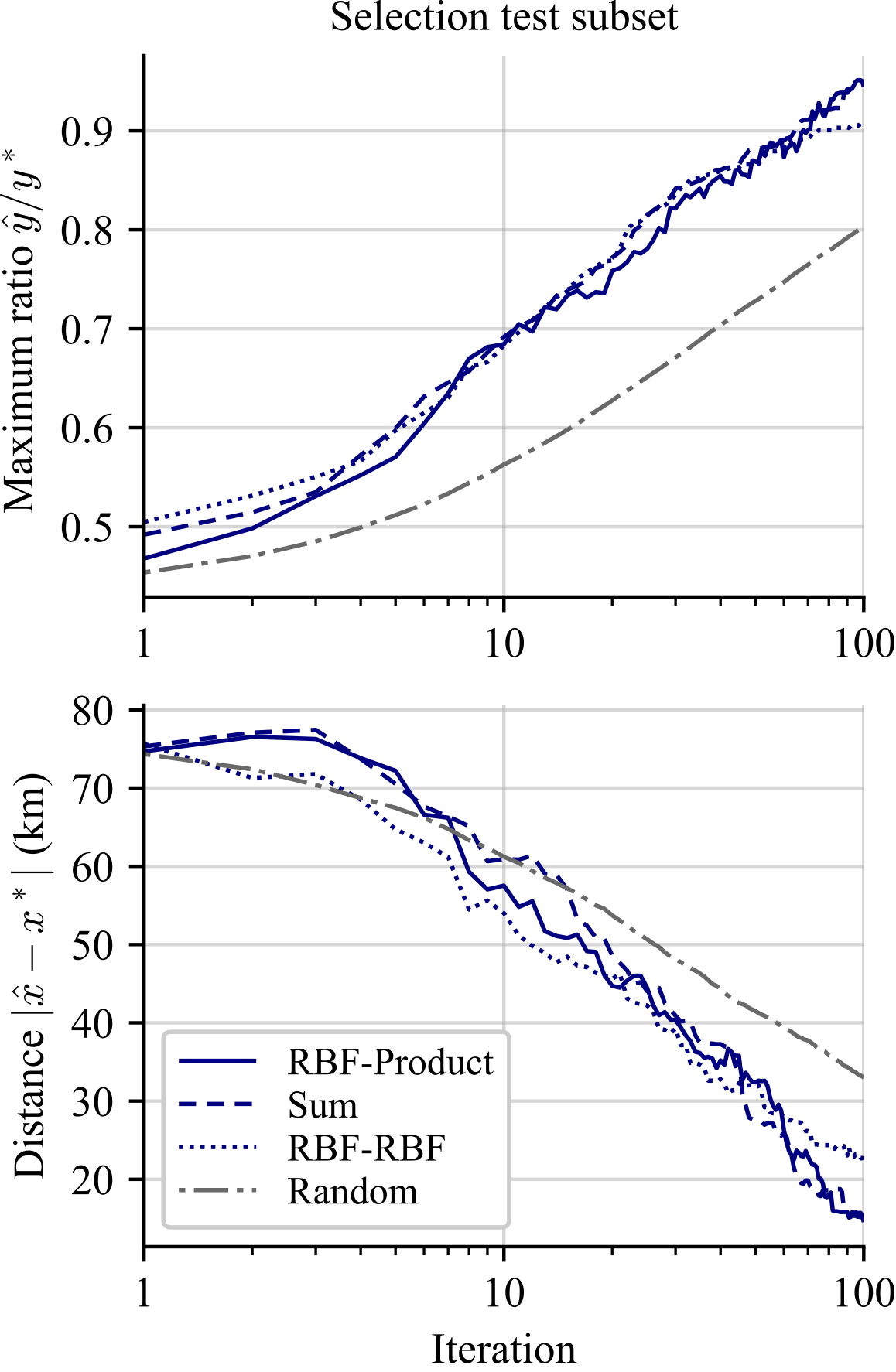}   
  \caption{Results on selection test subset of satellite data. 
  $\hat{x}$ is the estimated maximiser and $x^*$ the true maximiser. $\hat{y}$ and $y^*$ are the true concentration values at $\hat{x}$ and $x^*$, respectively.}
  \label{fig:satellite-selection-test-res}
\end{figure}

Additional results were acquired on a novel selection test subset, given in \cref{fig:satellite-selection-test-res,tab:satellite-selection-test-res}. The subset was created in the same way as the other selection subset, but without overlap. It provides additional evidence that BO is able to effectively locate the maximum.

\begin{table}[ht]
\centering
\caption{
Confidence intervals for the means of the maximum ratio and distance metrics at the final iteration for the selection test subset. We show means $\pm$ one standard deviation of the mean.
The best values are shown in \textbf{bold}. 
}
\label{tab:satellite-selection-test-res}
\begin{tabular}{@{}llll@{}}
\toprule
          & RBF-RBF     & Sum         & RBF-Product \\
Maximum\!\!    & \textbf{0.889-0.933}\!\! & \textbf{0.929-0.963}\!\! & \textbf{0.932-0.957} \\ \cmidrule(l){2-4} 
ratio          & Random       &        &   \\
          & 0.802-0.805  &  &  \\ \midrule
          & RBF-RBF     & Sum         & RBF-Product \\
Distance & \textbf{17.409-26.640}\!\!\! & \textbf{11.125-19.026}\!\!\! & \textbf{11.604-17.465} \\ \cmidrule(l){2-4} 
(km)          & Random       &        &   \\
          &  32.622-33.423\!\!\! &  &   \\ \bottomrule
\end{tabular}
\end{table}

\subsection{Comparison to earlier model}

\begin{figure}[ht]
  \centering
  \includegraphics[width= 0.45\textwidth]{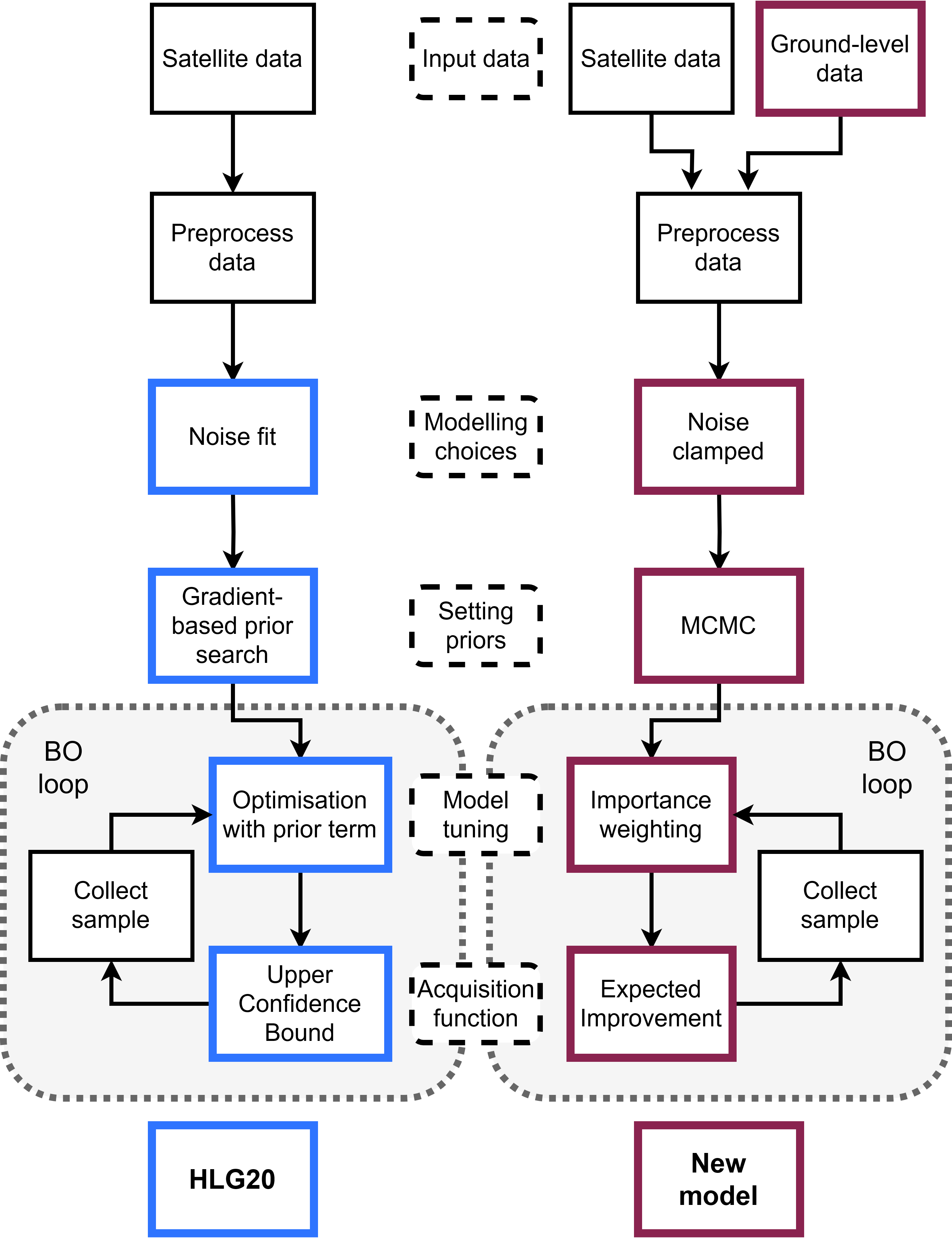}   
  \caption{Diagram of differences between model used in HLG20 (left) and our new model (right). The coloured boxes show where changes have been made. Only one of the satellite and ground-level data are used at the time, the figure is meant to show that the new work is evaluated on more data.}
  \label{fig:model_changes}
\end{figure}

\cref{fig:model_changes} shows the changes from the approach used in HLG20. 
The main differences are:   
\begin{itemize}
    \item[i)] Evaluation on more realistic data. In HLG20 only satellite data was used, while we here also use ground-level data.
    \item[ii)] Noise modelling. In HLG20, the noise hyperparameter was optimised, like the other hyperparameters. In the new model, the noise hyperparameter was instead clamped to a very low value.
    \item[iii)] Hierarchical modelling of data with MCMC for inference. HLG20 only used basic priors based on the maximum likelihood estimates (MLEs) on the tuning data. Instead, we here use a hierarchical model, obtaining samples of the hyperparameters, $\ta$, and of the parameters defining the distribution of the hyperparameters, $\ea$. 
    \item[iv)] Importance weighting instead of gradient-based optimisation. In HLG20, gradient-based optimisation is used to find the MLE on the test data at each iteration in the BO loop. Instead, we use importance weighting to fit our set of hyperparameter priors to the observed data at test time.
    \item[v)] Expected Improvement (EI) instead of Upper Confidence Bound (UCB). In HLG20, UCB was used as the acquisition function, but we found EI to perform better.
\end{itemize}

\subsection{Reproducibility}

\begin{itemize}
    \item \textbf{Data sets}: The London data set is freely available online \cite{imperial_college_london_london_1993_url}. The satellite data set is based on data freely available online at \url{https://scihub.copernicus.eu/}, but not the filtered version used in this work and the previous work compared to.
    \item \textbf{Seeds}: The random number generator used for MCMC is seeded with the number 13, as is also specified in the code. There is little randomness in the BO loop, since the noise is set very low.  
    \item \textbf{(Hyper-)parameter values}: The hyperparameter values needed to run the code are given, e.g. number of burn-in samples, number of BO iterations. The values for the model hyperparameters, e.g. the kernel lenghtscales for each model, are not given as they are outputs of the method and large in number.
    \item \textbf{Hyperparameter selection}: We do not have a full list of the hyperparameter settings tried. However, a general description of how they were set is given below. An exhaustive hyperparameter search to maximise performance was not conducted. The goal of the work was instead to evaluate the suitability of the method.
    \begin{itemize}
        \item \textbf{MCMC hyperparameters}: Different settings were tried for the number of samples and the proposal distribution. The number of samples was first tried at 100 or 200, but this was found to be too low for mixing. Instead, 1200 and 2000 were used for the satellite data and the London data, respectively. This was found to be a good balance between the required computation time and the usefulness of the samples. 
        The $\sigma_{\eta,k,g}$ values were experimented with to give good mixing. We first tried higher values but then reduced them to get a higher acceptance ratio. We found that higher values for the shape parameter $\psi$ than the scale parameter $\phi$ worked better.
        We initially tried updating all of $\theta$ jointly, and then all of $\eta$ jointly, but this led to much slower mixing.
        \item \textbf{Noise in likelihood}: We initially tried having the noise hyperparameter be learned, but found that this did not perform well. The models set the hyperparameter much higher, which led to a disconnect with the evaluation metrics as they assume the observed values to be ground truth.
        \item \textbf{GP models}: An exhaustive search of models was not done. We first tried the same models as in HLG20, (RBF, Sum and Product), but found that two kernels, one to model the slowly varying component and one for the faster varying component, worked better.
        \item \textbf{Acquisition functions}: We also tried probability of improvement and upper confidence bound, the latter setting the standard deviation scaling to 1, 4 and 10. We found that expected improvement worked best.
    \end{itemize}
    
\end{itemize}

\subsection{Execution times}

\begin{table}[ht]
\caption{Run times for maxima location (BO) and hyperparameter samples generation (MCMC). }
\begin{tabular}{@{}lllll@{}}
\toprule
                      & Kernel      & Strong & Selection & London \\ \midrule
\multirow{3}{*}{MCMC}   & RBF-RBF     & 38 min & 38 min    & 21 h   \\
                      & Sum         & 56 min & 56 min    & 30 h   \\
                      & RBF-Product & 71 min & 78 min    & 41 h   \\ \midrule
\multirow{3}{*}{BO} & RBF-RBF     & 44 min & 44 min    & 2 min  \\
                      & Sum         & 50 min & 52 min    & 2 min  \\
                      & RBF-Product & 49 min & 49 min    & 2 min  \\ \bottomrule
\end{tabular}
\label{tab:run_times}
\end{table}

The execution times are given in \cref{tab:run_times}. 
For MCMC, it is the time needed to collect the joint $(\ta,\ea)$ samples.
For BO, it is the average over data snapshots of the time needed for 100 iterations. 
For the London data, the MCMC takes between 21 and 41 hours, and the BO loops 2 minutes. For the satellite data, the MCMC takes between 38 and 78 minutes and the BO loops 44-52 minutes. 

BO is slower for the satellite data due to there being more candidate sampling locations. MCMC is slower for the London data due to there being more tuning data.
The MCMC time is less critical as it can be precomputed, while the BO time is encountered while placing sensors. The BO time is for 100 iterations, so even the slowest time, for the Sum kernel on the Selection subset -- 52 minutes --  averages to 31 seconds per iteration. This is negligible compared to the time required to seek out a location and install a sensor.
If necessary, the BO could be sped up by using posterior estimates for the mean and variance of the pollution level, instead of posterior estimates of the acquisition function.  
The BO complexity is linear in the number of hyperparameter samples from the prior.

\subsection{Computing infrastructure}

The experiments were run in parallel on nodes on a computer cluster running Scientific Linux 7.9 (Nitrogen). Each node had 14GB in memory and 8 CPUs. The CPU types used were:
\begin{itemize}
    \item Xeon CPU E5-2640 v3 @ 2.60 GHz
    \item Xeon CPU E5-2620 v3 @ 2.40 GHz
    \item Xeon CPU E5-2620 v4 @ 2.10 GHz
\end{itemize}
The code was run using Python 3.7.3. The complete Python dependencies are given in the code, but the main ones are NumPy (1.16.3) and SciPy (1.2.1).

\subsection{Data preprocessing}

The satellite data subsets were standardised using the means and standard deviations from the tuning sets. The LAQN data snapshots were mean-centred using data from that snapshot.
Both data sets were first log-transformed.

\end{document}